\documentclass[conference]{IEEEtran}
\usepackage{times}

\usepackage[numbers]{natbib}
\usepackage{multicol}
\usepackage{amsmath}
\usepackage{graphicx}
\usepackage{gensymb}
\usepackage{amsmath, xparse}
\usepackage{mathrsfs}
\usepackage{soul}
\usepackage{float}
\usepackage{xcolor}
\usepackage{stfloats}
\usepackage[bookmarks=true]{hyperref}
\usepackage{tikz}
\usetikzlibrary{matrix}
\usepackage{multirow}
\def\BibTeX{{\rm B\kern-.05em{\sc i\kern-.025em b}\kern-.08em
    T\kern-.1667em\lower.7ex\hbox{E}\kern-.125emX}}

\def\BibTeX{{\rm B\kern-.05em{\sc i\kern-.025em b}\kern-.08em
    T\kern-.1667em\lower.7ex\hbox{E}\kern-.125emX}}

\def\eg{{\em {\em e.g.},\ }}
\def\ie{{\em i.e.,\ }}

\begin{document}

\title{Adaptive Locomotion on Mud through Proprioceptive Sensing of Substrate Properties}

% You will get a Paper-ID when submitting a pdf file to the conference system
% \author{Author Names Omitted for Anonymous Review. Paper-ID [170]}

\thanks{
This research was supported by funding from the National Science Foundation (NSF) CAREER award \#2240075, and the NASA Planetary Science and Technology Through Analog Research (PSTAR) program, Award \# 80NSSC22K1313.}

% \thanks{$^{1}$Shipeng Liu, Jiaze Tang, Siyuan Meng and Feifei Qian are with Department of Electrical and Computer Engineering, University of Southern California, Los Angeles, CA, USA.
%         {\tt\footnotesize shipengl@usc.edu; jiazetan@usc.edu; siyuanm@usc.edu; feifeiqi@usc.edu}
%         (\textit{corresponding author: Feifei Qian})}%
%         }

\author{
  \authorblockN{Shipeng Liu\authorrefmark{2}, Jiaze Tang\authorrefmark{2},
                 Siyuan Meng\authorrefmark{2}, Feifei Qian\authorrefmark{2}\authorrefmark{1}}
  \authorblockA{\authorrefmark{1}Corresponding author}
  \authorblockA{\authorrefmark{2}Department of Electrical and Computer Engineering\\University of Southern California, Los Angeles, California, USA\\ Email: \{shipengl, jiazetan, siyuanm, feifeiqi\}@usc.edu}
}

% avoiding spaces at the end of the author lines is not a problem with
% conference papers because we don't use \thanks or \IEEEmembership

% \author{
% \authorblockN{Michael Shell\authorrefmark{1},
% Homer Simpson\authorrefmark{2},
% James Kirk\authorrefmark{3}, 
% Montgomery Scott\authorrefmark{3} and
% Eldon Tyrell\authorrefmark{4}}

% \authorblockA{\authorrefmark{1}School of Electrical and Computer Engineering\\
% Georgia Institute of Technology,
% Atlanta, Georgia 30332--0250\\ Email: mshell@ece.gatech.edu}
% \authorblockA{\authorrefmark{2}Twentieth Century Fox, Springfield, USA\\
% Email: homer@thesimpsons.com}
% \authorblockA{\authorrefmark{3}Starfleet Academy, San Francisco, California 96678-2391\\
% Telephone: (800) 555--1212, Fax: (888) 555--1212}
% \authorblockA{\authorrefmark{4}Tyrell Inc., 123 Replicant Street, Los Angeles, California 90210--4321}
% }

\maketitle
\begin{abstract}
Muddy terrains present significant challenges for terrestrial robots, as subtle changes in composition and water content can lead to large variations in substrate strength and force responses, causing the robot to slip or get stuck. This paper presents a method to estimate mud properties using proprioceptive sensing, enabling a flipper-driven robot to adapt its locomotion through muddy substrates of varying strength. First, we characterize mud reaction forces through actuator current and position signals from a statically mounted robotic flipper. We use the measured force to determine key coefficients that characterize intrinsic mud properties. The proprioceptively estimated coefficients match closely with measurements from a lab-grade load cell, validating the effectiveness of the proposed method. Next, we extend the method to a locomoting robot to estimate mud properties online as it crawls across different mud mixtures. Experimental data reveal that mud reaction forces depend sensitively on robot motion, requiring joint analysis of robot movement with proprioceptive force to determine mud properties correctly. Lastly, we deploy this method in a flipper-driven robot moving across muddy substrates of varying strengths, and demonstrate that the proposed method allows the robot to use the estimated mud properties to adapt its locomotion strategy, and successfully avoid locomotion failures. Our findings highlight the potential of proprioception-based terrain sensing to enhance robot mobility in complex, deformable natural environments, paving the way for more robust field exploration capabilities.

\end{abstract}

\IEEEpeerreviewmaketitle
\section{Introduction}
% \colorbox{green}{Feifei +1} % \colorbox{yellow}{Shipeng +2}\\

\begin{figure}[htbp]
  \centering
 \includegraphics[width=0.45\textwidth]{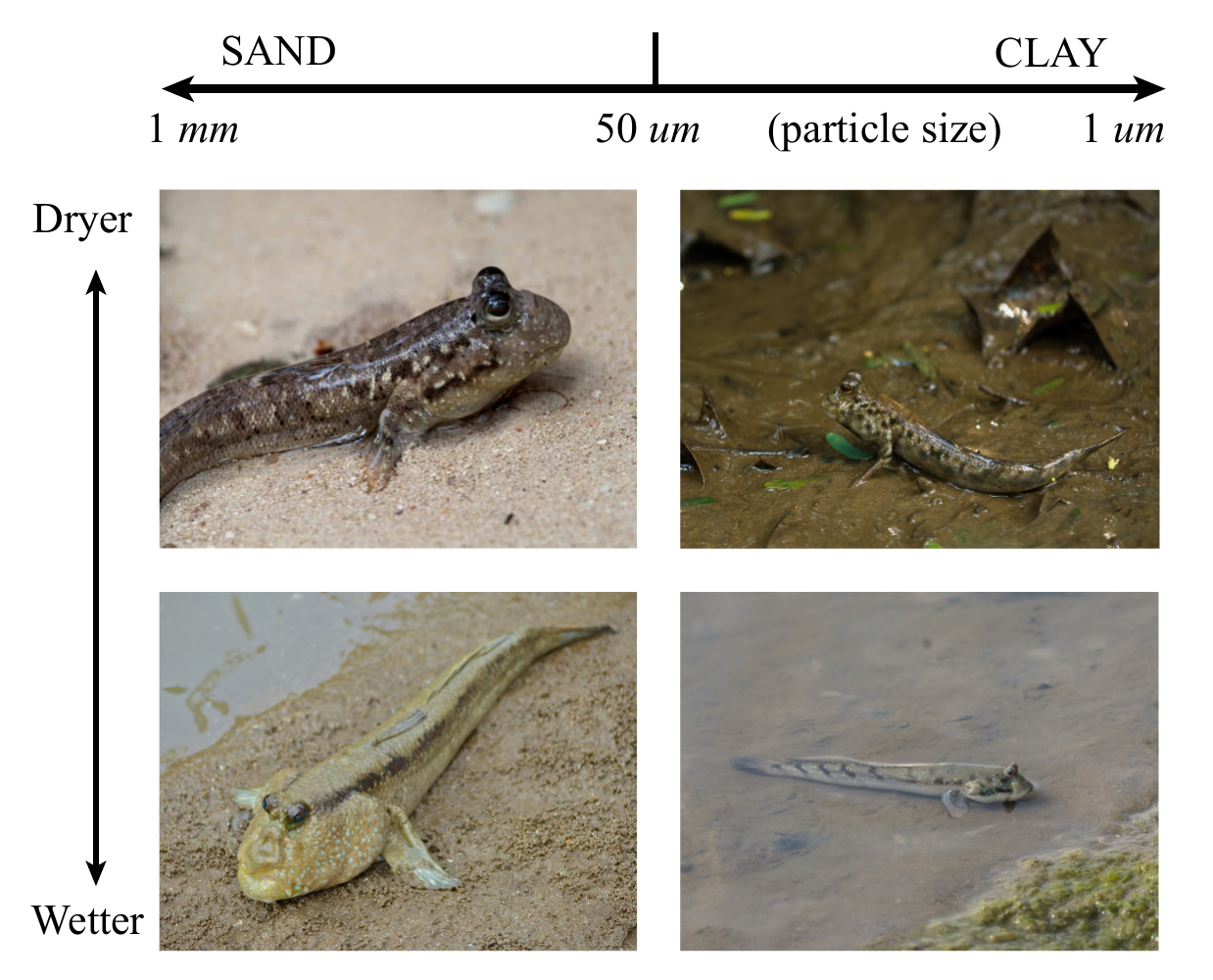}
  \caption{Natural mud mixtures vary significantly in water content and particle size. Inspired by mudskippers's ability to adapt locomotion strategies across a wide range of muddy substrates, this study explores bio-inspired strategies for proprioceptive sensing and locomotion adaptation on cohesive, deformable terrains. Photo courtesy Wikimedia Commons.}
  \label{Fig. mudskipper}
\end{figure}

Navigating natural deformable terrains can present significant challenges for terrestrial robots. This is because substrates like sand and mud can behave solid-like or fluid-like (Fig. \ref{Fig. mudskipper}) depending on their properties and how robots interact with them~\citep {shakeel2019rheological,nie2020investigation,coussot1994behavior,coussot1995structural,contreras2000coarse}. 
Previous studies have found that optimal locomotion strategies depend sensitively on substrate properties\citep{li2009sensitive, mazouchova2013flipper, qian2015principles, qian2013walking, liu2023adaptation}. Misapplied strategies, such as robot leg frequency or flipper speed, can lead to catastrophic locomotion failures. Successful movement on these substrates requires robots to accurately determine the force responses of the substrate~\citep{gravish2010force, maladen2011granular, qian2013walking, li2013terradynamics, gravish2014force} during their locomotion, and adapt their interaction actions accordingly. %Effective locomotion across such terrains requires adaptive strategies that respond to these dynamic changes.

%The complex interactions between the robot's flippers or legs and the terrain demand a detailed understanding of terrain dynamics to ensure robust adaptation. These interactions reflect varying mud properties and need to be sensed accurately during both static and dynamic movements~\citep{godon2023maneuvering}. 

Most existing studies on online locomotion adaptation target substrates of low deformability~\cite{marbach2005online,thor2019fast,kumar2021rma}, where the robot sinkage is relatively small as compared to the robot’s hip height.
% For example, \citet{marbach2005online} introduced a locomotion controller with coupled nonlinear oscillators and a genetic algorithm to enable rapid online gait adaptation after module failures. \citet{thor2019fast} implemented a dual integral learner for quick frequency adjustments in hexapod locomotion using neural CPGs and error feedback. \citet{kumar2021rma} train a terrain neural network in simulation to characterize terrain surface and adapt locomotion gait. 
However, how to effectively adapt robot locomotion on highly deformable complex terrains, such as fluidizing sand~\cite{qian2015principles} and watery mud~\cite{liu2023adaptation}, is largely unexplored.

Besides our limited understanding of the complex robot-substrate interactions~\cite{godon2022insight,godon2023maneuvering,liu2013understanding,chen2024reduced}, a key challenge for robot adaptation on sand and mud is due to the challenges in effectively measuring the properties of those substrates.  % -- the strength of dry sand can vary by 5 folds simply due to compaction, which is challenging to differentiate by vision.  
Existing terrain classification methods~\citep{8823987, 8794478, 10.3389/frobt.2022.887910, Fahmi2022OnTL, ren2024topnavleggednavigationintegrating} that rely on visual-based categorization often fail to capture the complexities of deformable terrain properties. Variations in terrain composition and compaction significantly influence robot locomotion performance, yet these factors are difficult to distinguish based solely on visual features.
%Many robotics researches~\citep{app10176044, QIN2023100091, 8793646,Fahmi2022OnTL} have already utilized various sensors, including vision~\citep{Fahmi2022OnTL,ren2024topnavleggednavigationintegrating}, lidar~\citep{8793646}, and tactile sensors~\citep{7397881, Bhattacharya2019SurfacePropertyRW}, to estimate terrain information. However, the above mentioned methods typically focus on classifying terrain based on known examples~\citep{app10176044,QIN2023100091} or estimate some general surface properties, such as friction and stiffness~\citep{Bhattacharya2019SurfacePropertyRW}, and lack the precision required for robots navigating complex deformable terrains~\citep{Bhattacharya2019SurfacePropertyRW}. For example, \citet{8793646} and \citet{Fahmi2022OnTL} developed distinct sensing methods, each using vision or depth information to estimate terrain properties. \citet{fu2022couplingvisionproprioceptionnavigation} coupled vision information with proprioceptive feedback, enabling their quadruped robot to sense terrain softness and slipperiness. Nevertheless, these methods above may not fully capture the dynamic and unpredictable nature of deformable environments, such as muddy terrain. 
%Calibrating terrain dynamics requires force information, as other sensory data like vision-based sensing often fail to capture the complex nature of deformable terrains with similar visual appearances. Although external f
Force sensors can provide valuable data for characterizing deformable substrate strength~\cite{7397881, Bhattacharya2019SurfacePropertyRW}, but those that are sensitive enough are often expensive and prone to damage under large impact forces during robot locomotion~\cite{7397881}, and resource-intensive in terms of integration and data acquisition~\citep{billeschou2021low, doi:10.1177/02783649211052067}. 

Recent research has demonstrated the potential of using direct-drive  motors~\citep{kenneally2018actuator} for proprioceptive terrain sensing. Given the reduced gear friction,  direct-drive motors offer high force transparency~\cite{Sungbae2016}, enabling terrain property measurement without the need for extra sensors~\cite{qian2019rapid}. However, most existing researches on substrate property characterization using direct-drive sensing were performed using single actuators or single robotic appendages mounted on a fixed frame~\cite{seok2012actuator, Sungbae2016, qian2019rapid, bush2023robotic}. How this method could be extended to characterize substrate properties during continuous robot locomotion, particularly over complex deformable terrains, remains unexplored. 

%While force estimation in static conditions has been demonstrated~\citep{seok2012actuator}, its accuracy in characterizing terrain dynamics during locomotion, particularly over deformable terrains, remains unexplored.

To address these challenges, in this study, we characterize the ability of a mudskipper-inspired robot to use proprioceptive joint signals to estimate ground reaction forces from synthetic mud mixtures (Sec. \ref{sec:motor_accuracy}). In addition to relating the ground reaction forces to interpret mud properties, we leverage existing granular physics models and develop a method to estimate mud properties from force signals measured from robot locomotion. To evaluate the accuracy of these estimations, we perform sensing and locomotion experiments with precisely-controlled and systematically-varied sand-clay-water ratios, and compare robot-estimated mud properties with the ground truth from standard load cells. Our result reveals the importance of considering the robot locomotion state jointly with ground reaction force for achieving an accurate interpretation of substrate properties (Sec. \ref{sec:online-sensing-results}). In addition, we show in Sec. \ref{sec:loco_adapt} that the proprioception-based measurements of mud properties can enable the robot to mitigate locomotion failures effectively identified in~\citep{liu2023adaptation} without relying heavily on external sensors. Such adaptability is crucial for field exploration robots, which often encounter diverse and unpredictable terrains~\citep{pace2009mudskipper, naylor2022mudskippers, you2014mudskipper}.

%first created various muddy substrates in the lab by mixing sand, clay, and water in different ratios~\citep{pradeep2023origins}. Subsequently, we validated the accuracy of proprioceptive sensing in measuring external forces as robots interact with the environment. Using a flipper-driven robot inspired by the mudskipper, we demonstrate how this proprioceptive sensing enables the online estimation of granular dynamics during flipper-mud interactions and validate the accuracy across static and dynamic movements, as detailed in Sec. \ref{sec:accuracy}. 

We summarize the main contributions of this paper as follows: 
\begin{itemize}
    \item Developed methods that allow robots to estimate mud penetration resistance, shear strength, and extraction resistive force from proprioceptive joint signals.
    \item Performed experiments and analysis to characterize the accuracy of the proposed method for estimating mud properties during continuous robot locomotion.
    \item Developed an online locomotion adaptation strategy that enables the robot to select locomotion strategies and mitigate locomotion failures based on online estimated mud properties.
\end{itemize}

\section{Materials and Methods}\label{sec: materials}
% \colorbox{yellow}{Shipeng +2}\\

% \colorbox{green}{Feifei +1} 
Using a mudskipper-inspired robot  (Sec. \ref{sec:robot}), we performed five sets of experiments, with specific focus on (1) characterizing the accuracy of proprioception-based torque measurements from individual actuators (Sec. \ref{sec:single-actuator-exp}); (2) characterizing and interpreting proprioceptive force signals from a statically-mounted robotic flipper to determine mud properties (Sec. \ref{sec:single-flipper-exp}); (3) relating proprioceptive force signals measured by a locomoting robot to determine mud properties (Sec. \ref{sec:robot-sensing-exp}); (4) comparing mud properties measured from robotic flippers with lab-grade load cell ground truth (Sec. \ref{sec:loadcell-exp}); and (5) demonstrating the ability of the robot to proprioceptively determine mud properties online, and adapt its locomotion to mitigate locomotion failures (Sec. \ref{sec:locomotion-exp}). 

\subsection{Robot}\label{sec:robot} %\subsubsection{Bio-inspired flipper-based robot}
% \colorbox{green}{Feifei +1}
The flipper-driven robot (Fig.~\ref{Fig. setup}A) used in this study was inspired by the mudskipper (\textit{Periophthalmus barbarus}~\cite{kawano2013propulsive} (Fig. \ref{Fig. mudskipper}), a small fish that can use its pectoral fins to produce effective locomotion on mud surfaces through a ``crutching'' motion. Modeled after the mudskipper's pectoral fins, two flipper arms (PLA plastic) were attached on the front side of the robot body (Fig. \ref{Fig. setup}A) to produce the flipper-driven locomotion on mud. The flipper actuators in this study were chosen to be direct-drive (\ie gearless) brushless DC motors (T-Motor, R60s). As compared to geared motors, direct-drive motors can offer high force transparency~\citep{kenneally2018actuator}, allowing the robot to estimate substrate reaction forces proprioceptively through joint position and joint current. 

The robot was programmed to use a mudskipper-inspired crunching gait~\cite{liu2023adaptation}, which is divided into four sequential phases: insertion, stance, extraction, and swing (Fig. \ref{Fig. setup}E). During the stance phase, the flipper maintains a constant insertion depth in mud, \( z \), by controlling the sweeping angle, $\alpha$, and adduction angle, $\beta$ (Fig. ~\ref{Fig. setup}B, C).

\subsection{Actuator sensing accuracy test}\label{sec:single-actuator-exp}

% \colorbox{green}{Feifei +1} 
The goal of this experiment is to characterize the accuracy of proprioception-based torque measurements from individual actuators. To do so, we compare the torque estimation from individual actuator with ground truth from calibration weights.

\begin{figure*}[htbp!]
  \centering
  \includegraphics[width=0.99
  \textwidth]{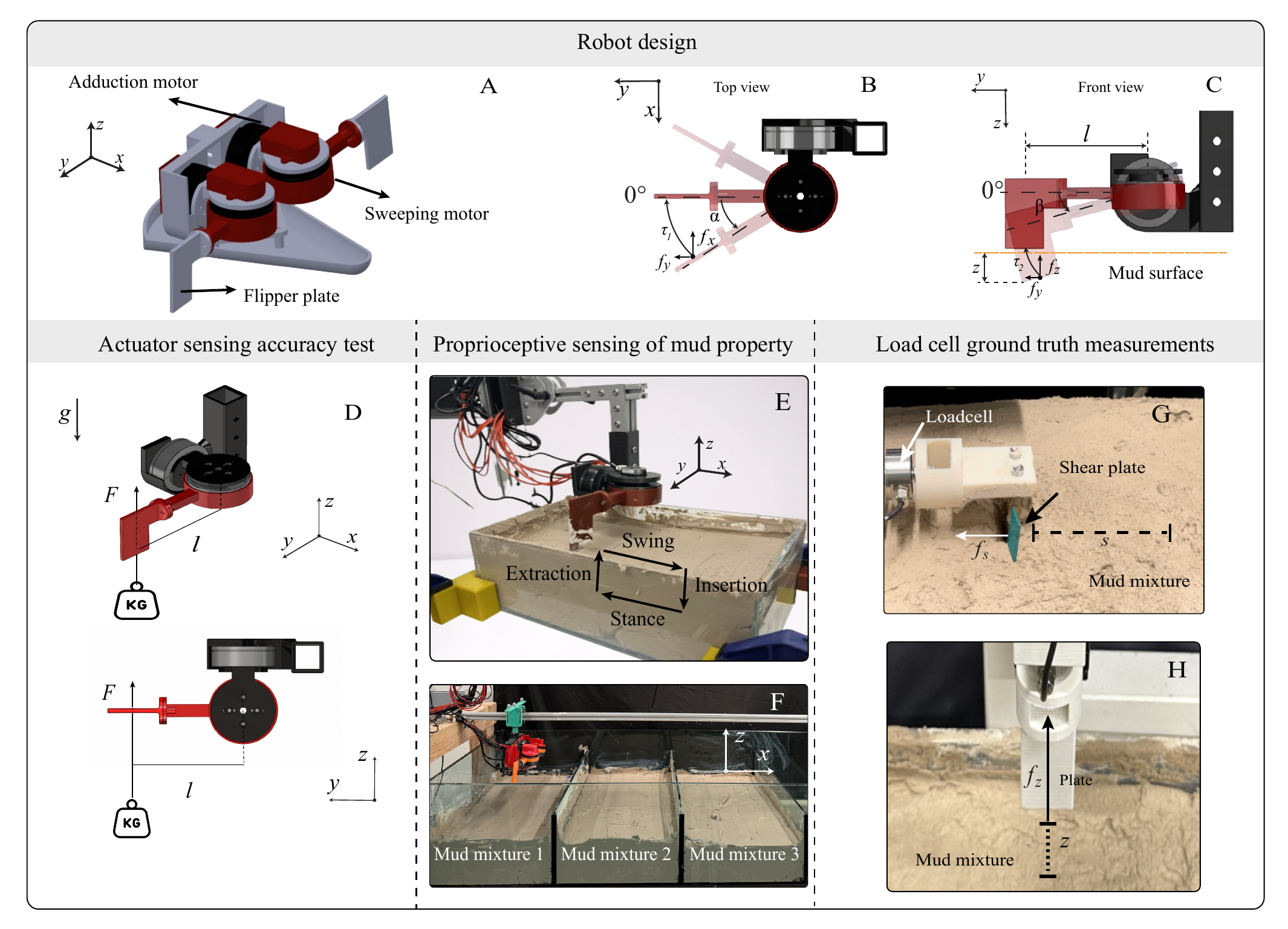}
  \caption{ Robots and Experiment Setup. (A) A bio-inspired flipper-based robot with a direct-drive motor. (B)(C) The flipper arm has two degrees of freedom (DoF): a sweeping angle, $\alpha$, and an adduction angle, $\beta$. The flipper arm length, l, is 11.5 cm. The flipper width and height are 2.5 cm and 7 cm, respectively. (D) Single flipper motor calibration experiment setup. (E) A single flipper robot force experiment setup, with a fixed base, tests flipper-mud interaction forces in a fixed glass tank. (F) Locomotion experiment setup: A four-segment trackway allows robot traversal through different mixtures in a single trial, with two motion capture cameras tracking the robot’s fore-aft position, while a supporting rail controls the flipper insertion depth precisely.  (G)(H) Linear actuator setup for horizontal shear and vertical insertion/extraction experiments.    }
  \label{Fig. setup}
\end{figure*}

We isolated one robotic flipper and mounted it in two configurations. In the first configuration (Fig. \ref{Fig. setup}D, top diagram), the axle of the adduction motor was kept horizontal, and a calibration weight was suspended beneath the flipper, with a horizontal distance $l$ from the adduction motor axle. 
In the second configuration (Fig. \ref{Fig. setup}D, bottom diagram), the axle of the sweeping motor axle was kept horizontal, and a calibration weight was suspended beneath the flipper with a horizontal distance $l$ from the sweeping motor axle. 

Both motors were programmed to hold constant positions during the trial, $\ alpha=0$ and $\ beta=0$. The proprioceptively estimated external torque ~\citep{seok2012actuator, Sungbae2016} was computed as $\tau_{\text{sense}} \approx  k_t I$, where $I$ is motor current, and $k_t = 0.083$ Nm/A is the motor torque constant~\footnote{The damping friction force and torque-dependent Coulomb friction force are negligible~\citep{kenneally2018actuator} and therefore not included in the computation.}. The ground truth of the external torque, $\tau_{\text{ext}}$, can be computed as the gravitational force, $mg$, of the suspended load, multiplied by the moment arm, $l$. Data were collected across the torque range of the selected motor, with suspended load of 10 g, 20 g, 50 g, 100 g, 200 g, 400 g, 600 g, and 800 g. Five trials were collected per calibration weight for both configurations.

\subsection{Single flipper measurements of mud property}\label{sec:single-flipper-exp}

% \colorbox{green}{Feifei +1} 
The goal of this experiment is to characterize mud resistive forces proprioceptively from an individual robotic flipper and determine how these force signals can be interpreted to determine mud properties.

To characterize the mud resistive forces, we mounted the single robotic flipper above a container of synthetic mud (Fig.~\ref{Fig. setup}E). The synthetic mud used in this study was created by mixing sand (Pottery Supply House, 150–600 {$\rm \mu m$} Silica Sand), clay (Seattle Pottery Supply, Edgar Plastic Kaolin), and water. The sand-clay-water mixture has been shown to produce similar rheological behaviors of natural muds~\citep{coussot1994behavior,kostynick2022rheology}, while allowing for systematic variation of its properties. In this study, we kept the clay-to-sand ratio constant, 3:1, and systematically adjusted the water content, $w$, to control mud strength. In this experiment, $w$ was varied at five levels, 47.6\%, 48.6\%, 49.5\%, 50.3\%, and 51.2\%. This range was selected between the fracture limit and the settling limit~\cite{coussot2017mudflow}, where most robot locomotion failures occur~\cite{liu2023adaptation}. Below the fracture limit, the mud would behave like fractured solid and cannot be fully mixed; whereas beyond the settling limit, the mud is fully saturated and solid and liquid would begin to separate. 
% {\FQ note to self: this is 1 - $\phi$, computed as water volume / (solid mass/density + water volume), which differs from the RAL paper, which was water volume / (water + occupied solid volume before mixing).}

During each trial, the robotic flipper was programmed to execute one cycle of the crunching gait (Fig.~\ref{Fig. setup}E, Insertion, Stance, Extraction, and Swing phases). In this set of experiments, the linear velocity of the flipper was set to 0.1 m/s during all phases, and a three-second pause was added between the phases to leave time for force relaxation. The insertion depth, $z$, was set to be 3 cm for this set of experiments. For all trials in this study, the sweeping range during the stance phase was kept at \( [-30, 30] \) degrees. Motor angular position and current were logged at 380 Hz. The external forces exerted on the robot flipper, $\mathbf{F}_{\text{sense}} = [f_x, f_y, f_z]^T$ (Fig. \ref{Fig. setup}B, C), were computed as $\mathbf{F}_{\text{sense}} \approx  \mathbf{J}^\mathrm{-T} \mathbf{\tau}_{\text{sense}}$, where $\mathbf{J}$ is the Jacobian of the robotic flipper, and $\mathbf{\tau}_{\text{sense}} = [\tau_1, \tau_2]^T$ is the proprioceptively estimated motor torque. During the robot stance phase, the lateral forces, $f_y$, from both flippers cancel each other out, and we computed the vertical force \( f_z \) and the fore-aft force \( f_x \) for estimating the terrain coefficients:\[
\begin{bmatrix}
f_x \\
f_z
\end{bmatrix}
=
\begin{bmatrix}
\frac{\cos \alpha}{l} & 0 \\
-\frac{\sin \beta \cdot \sin \alpha}{l} & \frac{\cos \beta \cdot \cos \alpha}{l}
\end{bmatrix}
\begin{bmatrix}
\tau_1 \\
\tau_2
\end{bmatrix}
\]

\subsection{Online mud property mapping experiments}\label{sec:robot-sensing-exp}

% \colorbox{green}{Feifei +1}

The goal of this experiment is to determine mud properties through proprioceptive joint signals from a locomoting, flipper-driven robot.

The experiments were performed in a modular mud trackway (183 cm $\times$ 76 cm $\times$ 30 cm) (Fig. \ref{Fig. setup}F). The two removable panels of the trackway allow mixtures in each separable compartment to be prepared with different properties (\eg, water content, or solid compositions). Once the mixtures are prepared, panels can be lifted, allowing the robot to locomote across the entire length of the trackway. This testbed enables robot sensing and locomotion experiments on transitional terrain properties. In this set of experiments, we prepared the mud mixture in the three compartments with water content $w$ = 47.6\%, 49.5\%, and 51.2\%. 

% Previous study has shown that this range of water content resulted in distinct mud strength and rheology, leading to different robot locomotion failure modes~\cite{liu2023adaptation}.

A mudskipper-inspired robot was mounted on a linear rail, and programmed to move through the mud trackway. The robot used the same gait parameters as described in Sec. \ref{sec:single-flipper-exp}. During each step, the robot logs its motor position and current at 380 Hz, and estimates mud properties using the proprioceptive signals from every step. The projected surface area and insertion depth of the flipper in mud were computed from the angular positions of the motors, measured by encoders. Three trials were performed for this experiment. Before each trial, the surface layer of the mud mixture was manually re-mixed and leveled to ensure consistency between trials. Two Optitrack Prime 13W motion capture cameras recorded the robot's position in the world frame ($x$, $y$, $z$) at 120 FPS. An additional webcam was used to capture zoomed-in side-view videos at 30 FPS, focusing on flipper-mud interactions and robot footprints.

\subsection{Load cell ground truth measurements}\label{sec:loadcell-exp}
% \colorbox{yellow}{Shipeng +1}
The goal of this experiment is to provide the ground truth for robot-estimated mud properties in Sec. \ref{sec:single-flipper-exp} and Sec. \ref{sec:robot-sensing-exp} and evaluate the accuracy. 

To determine the ground truth of mud coefficients, we employed a high-speed linear actuator (Heechoo) to actuate a flipper plate matching the size of the robot’s flipper. Two sets of force measurements were conducted: (i) horizontal shear (Fig. \ref{Fig. setup}G), where the robotic flipper plate (3 × 3.5 cm) was dragged horizontally across the mud at a constant insertion depth of 3 cm, covering a shear distance of 20 cm; and (ii) vertical insertion and extraction (Fig. \ref{Fig. setup}H), where the flipper plate was inserted vertically into the mixture to a depth of 3 cm, held stationary for 3 seconds, and then extracted. The force responses were measured using a loadcell (DYMH-103) mounted between the flipper and the
linear actuator. The shear, insertion, and extraction velocities were set to  0.1 m/s, the same as the single flipper experiments in Sec. \ref{sec:single-flipper-exp}.  

We conducted loadcell experiments on mud mixtures with water content, 47.6\%, 48.6\%, 49.5\%, 50.3\%, and 51.2\% to provide a comparison with results from Sec. \ref{sec:single-flipper-exp} and Sec. \ref{sec:robot-sensing-exp}. For each water content, ten trials were performed, five for horizontal shear, and five for vertical penetration and extraction. 

\subsection{Locomotion adaptation experiments}\label{sec:locomotion-exp}
% \colorbox{yellow}{Shipeng +1}
The goal of this experiment is to demonstrate that by proprioceptively determining mud properties online, a flipper-driven robot would be able to flexibly adapt its locomotion and mitigate locomotion failures.

To do so, we evaluated the robot locomotion performance in the modular mud trackway, crossing transitioning mud properties with water content 48.5\%, 53.8\%, 45.9\%, respectively (Fig. \ref{Fig. setup}F). A previous study~\cite{liu2023adaptation} has suggested that this range of water content variation can result in distinct mud strength and rheology, leading to different locomotion failure modes with misapplied robotic flipper actions. Specifically, mud with higher water content (\eg 53.8\%) exhibited low yield strength and could cause flipper slippage with small flipper insertion depth, whereas mud with lower water content (\eg 45.9\%) exhibited high cohesion and could lead to flipper entrapment with large flipper insertion depth. 

%The left segment, with an intermediate water content of 48.5\%, posed a minimal risk of failure. The middle segment had a high water content of 53.8\%, resulting in insufficient yield stress to support forward movement. The right segment, with a low water content of 45.9\%, created a highly cohesive mixture that hindered flipper extraction if the insertion depth exceeded a critical threshold.

Based on this, We tested three flipper insertion depth across the transitioning muddy terrain: (1) a constant 3 cm insertion depth across all three mud mixtures; (2) a constant 5 cm insertion depth across all three mud mixtures; and (3) an adaptive insertion depth based on proprioceptively-estimated mud properties. Three trials were performed for each configuration. 
The mud surface was manually flattened between the trials. 
Flipper velocity was set to 0.1 m/s during the insertion phase to enhance sensing accuracy, and kept at 0.5 m/s during the Stance, extraction, and swing phases to enable faster locomotion. No additional pauses were added between the phases. %{\FQ this information should also be reported for Sec. \ref{sec:robot-sensing-exp}.}
During each trial, the commanded insertion depth, robot motor position and current, and tracked robot pose were recorded at 380 Hz.

\section{Results}\label{sec:results}
% \colorbox{yellow}{Shipeng +2}\\
% \colorbox{yellow}{Shipeng +1}
In this section, we first report the characterized accuracy of torque estimation from each joint actuator (Sec. \ref{sec:motor_accuracy}). We then show proprioceptive force signals measured using a statically mounted robotic flipper, and report our method to estimate mud properties from these signals (Sec. \ref{sec:forcemodel}). These proprioceptively estimated mud properties are compared to measurements by a load cell to quantify the accuracy.
In Sec. \ref{sec:online-sensing-results}, we extend the proposed mud property estimation method to a locomoting robot and investigate the additional challenges introduced by body movements. Finally, we demonstrate the ability of our robot to proprioceptively determine mud properties online and adapt its locomotion to mitigate locomotion failures (Sec. \ref{sec:loco_adapt}).

\subsection{Torque sensing accuracy characterized from single actuator experiments}\label{sec:motor_accuracy} %Motor transparency validation
\begin{figure}[htbp]
  \centering
  \includegraphics[width=0.42\textwidth]{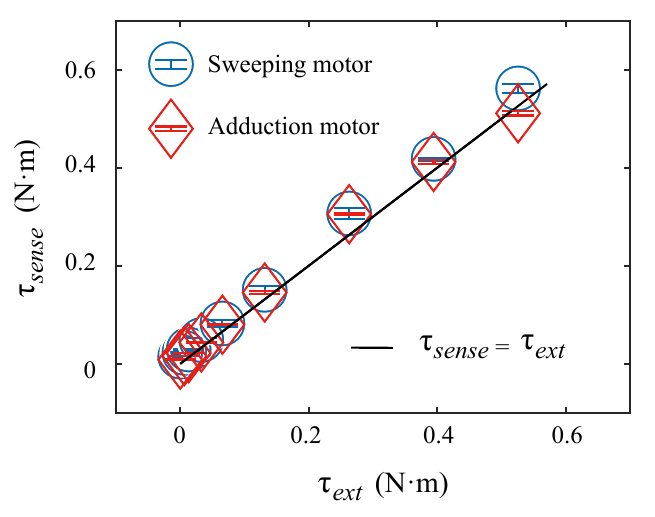}
   \caption{Comparison between the proprioceptively-estimated torque, $\tau_{sense}$, and the ground truth external torque based on calibration weights, $\tau_{ext}$.}
   
   % (C) The force measurement from the flipper motion and transformed force through Jacobian. (D) The relationship between the vertical force, $F_z$, and insertion depth, $d$. (E) The relationship between the sweeping force, $f_s$, and sweeping angle, $\alpha$. (F) The estimated insertion stiffness, $k_i$, average shear force, $\hat{f_s} $, and maximum extraction force, $\hat{f_e}$ of different water content, $W$, increasing from xxx(blue) to xxx(cyan){\SL consider changing a color map}. (G) Estimation results comparison between robot measurements and linear actuator measurements.}
  \label{Fig. sensing_validation}
\end{figure}

% \colorbox{yellow}{Shipeng +2}
Using the experimental setup described in Sec. \ref{sec:single-actuator-exp}, we quantified the accuracy of torque estimation by comparing the proprioceptively-estimated torque, $\tau_{sense}$ (Fig.~\ref{Fig. sensing_validation}, y-axis), with the ground truth external torque, $\tau_{ext}$ (Fig.~\ref{Fig. sensing_validation}, x-axis). The external torque, $\tau_{ext}$, for both the adduction motor (Fig.~\ref{Fig. sensing_validation}, blue circles) and the sweeping motor (Fig.~\ref{Fig. sensing_validation}, red diamonds) closely aligns with the ground truth, with a RMSE (Root Mean Square Error) of 0.0134 N/m, and 0.0127 N/m, respectively.
% , calculated as $e = (\tau_{sense} - \tau_{ext})$.{\SM{The average error percentage is approximately 48\%, calculated as $e = (\tau_{sense} - \tau_{ext}) / \tau_{ext}$.}}
Furthermore, the standard deviation of $\tau_{sense}$ remains very low, with an average of 3.2\% related to the $\tau_{sense}$, indicating a good repeatability of the torque estimation.
We note that all torque estimations were produced using motor current and the manufacturer-provided torque constant without requiring any fitting parameters. 
These results demonstrated the torque sensing accuracy of direct-drive actuators, and confirmed the validity of the proposed method of using motor current to estimate external forces in the subsequent sections. 

\subsection{Determining mud properties from proprioceptively-sensed mud reaction forces}\label{sec:forcemodel}
% \colorbox{yellow}{Shipeng +1}

\begin{figure}[htbp]
   \centering  \includegraphics[width=0.5\textwidth]{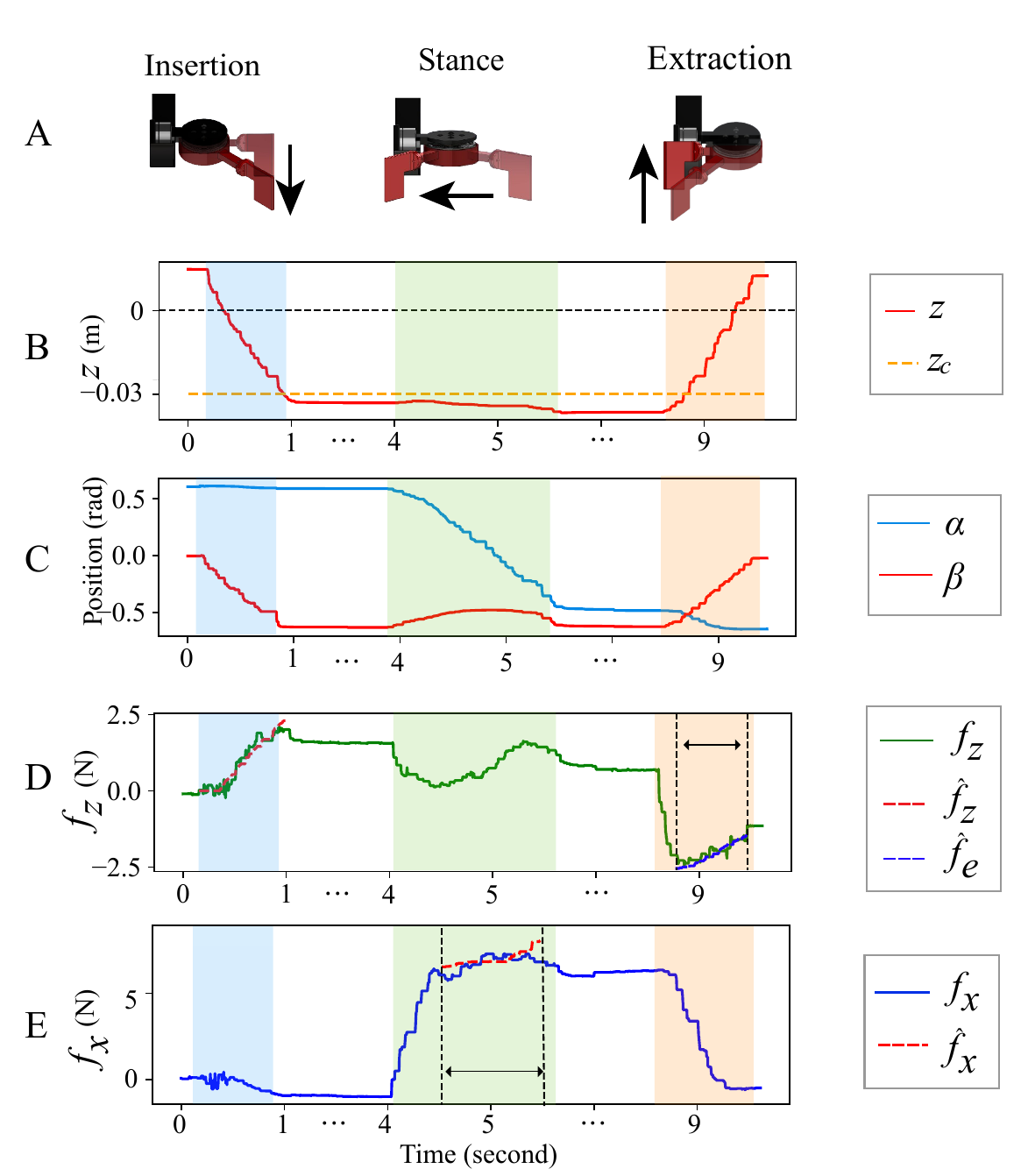}
  \caption{Measurements from the single flipper experiment.  (A) During one full gait cycle, the robotic flipper executes three phases of movements that produce distinct mud force responses: insertion, stance, and extraction. (B) The commanded insertion depth \(z_c\) and actual insertion depth \(z\) versus time in one gait cycle.  (C) The recorded actuator position for the sweeping motor (blue curve) and adduction motor (red curve) within one gait cycle. (D) The proprioceptively-measured vertical external force, \(f_z\) (green curve), within one gait cycle. The modeled penetration force, $\hat{f_z}$, and the modeled extraction force, \(\hat{f_{e}}\), are represented as dashed red curve and dashed dark blue curve, respectively.  (E) The proprioceptively-measured horizontal external force, \(f_x\) (blue curve), within one gait cycle. The modeled yielding force, $\hat{f_x}$, is represented as the dashed red curve. The blue, green, and orange shaded regions in B-E correspond to the flipper insertion, stance, and extraction phases, respectively.}
  \label{Fig. sensingFig1}
\end{figure}

Here, we report our method for determining intrinsic mud properties from the proprioceptive joint signals. Using the experimental setup described in Sec. \ref{sec:single-flipper-exp}, we measured actuator-estimated external torque for flipper motors, as the stationary-mounted robotic flipper performs a full cycle of its crunching gait (Fig. \ref{Fig. sensingFig1}A).  
Propagating the actuator-estimated external torque through the flipper kinematics allows us to determine the external force, $F_{sense}$, acting on the flipper during its interactions with muddy terrains (Fig. \ref{Fig. sensingFig1}D, E). %To leverage the rich information contained in the force profile during the interaction, we need a model to represent and interpret the force profile and characterize the robot-mud interactions effectively. 
In this section, we use the proprioceptively measured forces to determine mud properties by leveraging granular physics force principles.

%Below we discuss the proprioceptively-measured force responses, and how to connect them to intrinsic mud properties.

% By extracting these coefficients from the corresponding movement, we can characterize the terrain and leverage this information to adapt gaits and prevent locomotion failures online. Details of each coefficient calculation are presented as follows. 

%\subsubsection{Determine robot gait trajectory for sensing deformable terrain strength}

\subsubsection{Determine mud penetration resistance}\label{sec:kp}

During the insertion phase (Fig. \ref{Fig. sensingFig1}, blue shaded area), the proprioceptively-measured $f_z$  exhibited a linear increase as the flipper inserted into the terrain (Fig. \ref{Fig. sensingFig1}D, green solid curve). This is because the resistance force of substrates like sand and mud on penetrating intruders is primarily governed by the hydrostatic-like pressure~\citep{kang2018archimedes}. Previous research in granular physics~\cite{albert1999slow,katsuragi2007unified} has found that this pressure scales linearly with insertion depth, \( z \) (Fig. \ref{Fig. setup}C), the projected intruder surface area perpendicular to the motion, $A$, and the substrate's penetration resistance, $k_p$.

\begin{figure*}[htbp]
  \centering
  \includegraphics[width=1\textwidth]{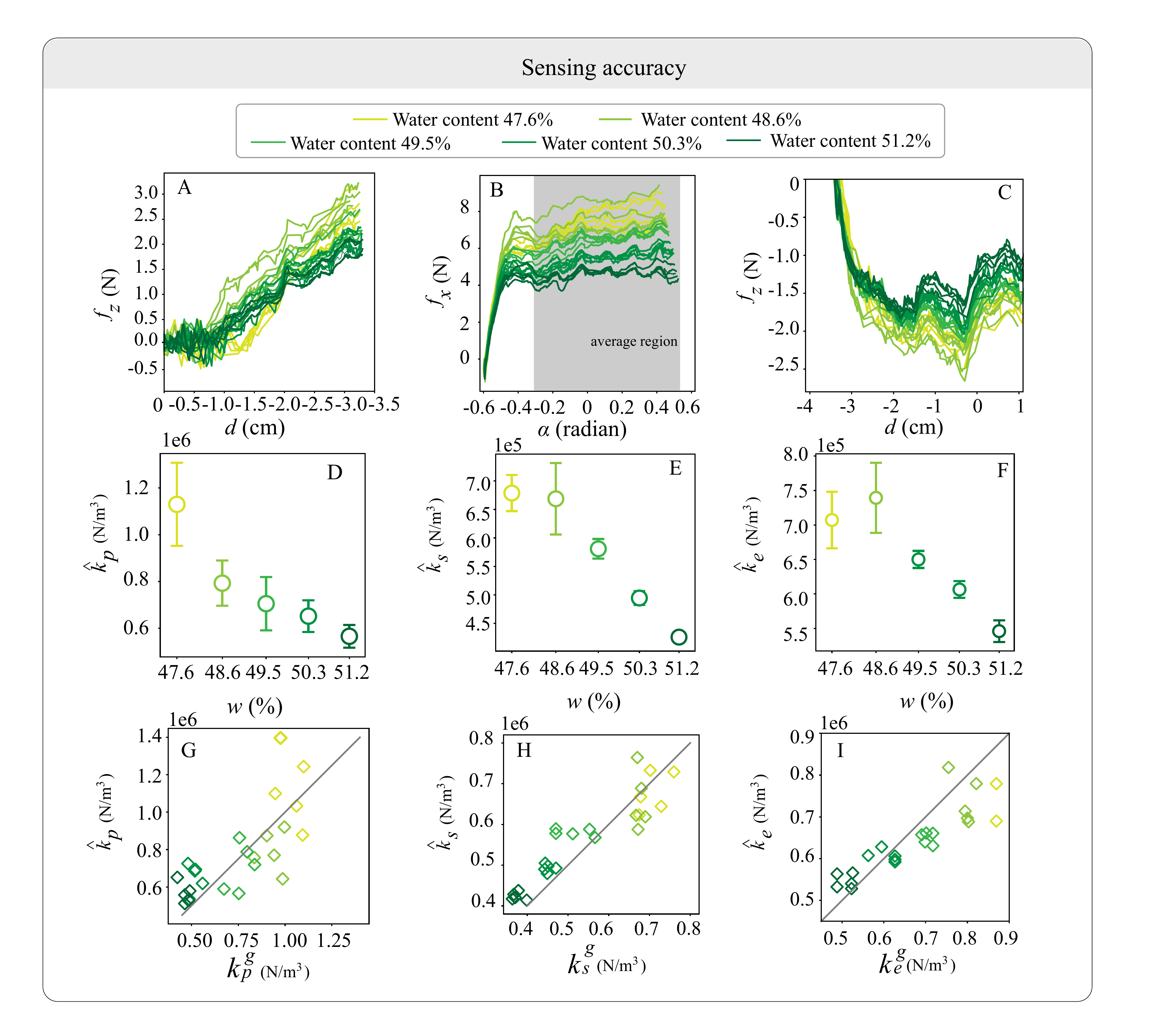}
  \caption{Validation of robot-sensed mud properties. \textbf (A) Cropped $f_z$ versus insertion depth, $z$ in the insertion phase. (B) Cropped $f_x$ versus sweeping angle, $\alpha$ in sweeping phase. (C) Cropped $f_z$ versus insertion depth, $z$, in the extraction phase. (D)(E)(F) Mud properties, $\hat{k}_p$, $\hat{k}_s$ and $\hat{k}_e$ sensed by robot. (G)(H)(I) Comparison between robot-sensed mud properties, $\hat{k}_p$, $\hat{k}_s$ and $\hat{k}_e$, and the corresponding ground truth measured by the loadcell. The grey line indicates that the estimated results are equal to the ground truth. For all plots, mixtures with different water content are represented by different colors.}
  \label{Fig. sensing_features}
\end{figure*}

For dry, deformable granular terrains like sand, the penetration resistance has been a widely used metric to evaluate the substrate strength~\citep{stone2004local,li2009sensitive,qian2015principles}. Recent studies in rheology and robot locomotion~\citep{godon2022insight, liu2023adaptation, chen2024reduced} have shown that despite the significantly increased cohesion and more complex rheological responses~\cite{shakeel2019rheological,nie2020investigation,coussot1998direct,contreras2000coarse,coussot1994behavior,jerolmack2019viewing} of muddy substrates, penetration resistance remains an effective measure to characterize the strength of these materials. Based on these findings, here we seek to determine penetration resistance as one of the key parameters to quantify mud properties.

One challenge in determining $k_p$ using a locomoting robot appendage is that, based on flipper motion, its insertion depth in the substrate, $z(t)$, and project surface area, $A(t)$, could vary with time, resulting in challenges in interpreting the resulting force responses and extracting $k_p$. 
To address this challenge, we leverage the granular resistive force theory (RFT)~\cite{li2013terradynamics}. According to RFT, the granular resistive force on an object moving through a granular medium can be computed as the integral of infinitesimal resistive forces from segments at different depths. For each segment, the force was a function of the segment's orientation and moving direction. By summing up these infinitesimal forces across the entire object (\eg, the robotic flipper), the theory gives an accurate prediction of the total resistive force that the object experiences in granular media. While RFT has been originally developed for dry granular media, recent work has demonstrated its applicability to be used for cohesive materials~\cite{kerimoglu2024extending}.

Based on RFT, we determine $k_p$ by minimizing the RMSE between the modeled $\hat{f_z}(k_p)$ using Eqn. \ref{eqn1} and actual measurements, $f_z$:  
\[
k_p = \operatorname*{arg\,min}_{k_p} \left( \hat{f}_z(k_p) - f_z \right)^2
\] 
Here $\hat{f_z}(k_p)$ was computed based on RFT, by integrating the penetration force from each infinitesimal flipper bottom segment at different depth \( z \) (Fig. \ref{Fig. setup}C):
\begin{equation}
     \hat{f_z} = k_p \int A(z) \, dz 
    \label{eqn1}
\end{equation}
where \( d\hat{f_z} = k_p \cdot A(z) \cdot dz\) is the force exerted on each infinitesimal segment at \( z \).
Since $k_p$ is an intrinsic mud property and does not depend on flipper motion, as the flipper moves through the mud and gathers data of $f_z$, it can use these data to estimate $k_p$.

%\( A(z) \) is the projected surface area of the caisson {\FQ flipper?} at depth \( z \), and \( \hat{f_z} = k_p \int A(z) \, dz \) at each moment during the insertion phase. 

%Then, we use the list of estimated $f_z$ (blue shaded area in Fig. \ref{Fig. sensingFig1}D) and pre-calculated \( \int A(z) \, dz \) during the insertion phase to estimate the $k_p$ by minimizing the RMSE error between the modeled $\hat{f_z}(k_p)$ using Eqn. \ref{eqn1} and actual measurements, $f_z$  \[k_p = \operatorname*{arg\,min}_{k_p} \left( \hat{f}_z(k_p) - f_z \right)^2\] 

Fig. \ref{Fig. sensing_features}A shows the proprioceptively measured $f_z$ by the robotic flipper during the intrusion phase. The force profiles exhibit similar patterns across different water content, but the force magnitude decreased monotonically as the water content increased from 47.6\% to 51.2\% (Fig. \ref{Fig. sensing_features}A, light to dark green).
Fig. \ref{Fig. sensing_features}D plots the estimated mud penetration resistance, $k_p$, for each water content, using the proposed method, 
The estimated $k_p$ was inversely related to the water content, which is consistent with previous literature~\cite{liu2023adaptation}.
Using the estimated $k_p$, we plotted the modeled $\hat{f_z}$ in Fig. \ref{Fig. sensingFig1}D, red dashed line in blue shaded region, which aligns well with the measured $f_z$ (Fig. \ref{Fig. sensingFig1}D, green curve in blue shaded region), and the ground truth measurements from the load cell (Fig. \ref{Fig. sensing_features}G).

% We note that the insertion motion will also result in a $f_y$ force. However, we do not consider using the shear force in the y direction to estimate the coefficients because the flipper has a very short travel distance along the y direction. 
\subsubsection{Determine mud shear strength} \label{sec:ks}
% {\FQ first describe force measurements observed during the stance phase (Fig. \ref{Fig. sensingFig1}D). Then discuss previous literature, to provide a rationale on why and how we compute shear strength to represent mud property.}

The horizontal mud resistive force, $f_x$, remains near zero during the insertion phase (Fig. \ref{Fig. sensingFig1}E, blue shaded area), but began to increase rapidly to a steady value as the flipper shears through mud during the stance phase (Fig. \ref{Fig. sensingFig1}E, green shaded area). 
This is consistent with previous granular literature, which has found that the shear resistance force, \( \hat{f_x} \), exerted on an object moving horizontally, scales linearly with the shear depth \( z \), the shear object's projected area perpendicular to the motion direction \( A \), and the material's shear strength coefficient, \( k_s \)~\citep{albert1999slow,li2013terradynamics,qian2019rapid}. %Here we seek to determine mud shear strength, $k_x$, as one of the key parameters to quantify mud property.

%For horizontal movements, we model shear strength using the Mohr-Coulomb model~\citep{punmia2005soil}, which assumes purely plastic behavior of mud mixture as the applied force is larger than the yielding strength. Thus, as the mud yields, the instantaneous horizontal shear force, \( f_x \), equals the modeled yielding force, \( \hat{f_x} \).

%{\FQ note to self: differentiate shear force and shear strength}
Based on the granular force principle, we estimate \( k_s \), %from the proprioceptively measured force (Fig. \ref{Fig. sensingFig1}D) and insertion depth (Fig. \ref{Fig. sensingFig1}B), 
by minimizing the RMSE between the modeled shear resistive force, $\hat{f_x}(k_s)$, and proprioceptively-sensed shear resistive force, $f_x$, during the steady-state region (Fig. \ref{Fig. sensingFig1}E, between vertical dashed lines): 

\begin{equation}
    k_s = \operatorname*{arg\,min}_{k_s} \left( \hat{f_x}(k_s) - f_x \right)^2
    \label{eqn3}
\end{equation}

Here $\hat{f_x} = k_s \int bz \, dz$ is an integral of the shear force from each submerge infinitesimal flipper segments at depth $z$ (Fig. \ref{Fig. setup}C), %, where $d\hat{f}_x = b \cdot k_s \cdot z \cdot dz$ is the shear resistive force exerted on each infinitesimal segment at depth $z$ (Fig. \ref{Fig. setup}C). 
where \( b \) is the width of the segment, \( b \cdot dz \) represents the projected surface area of the flipper segment along the sweeping direction. %, and \( k_s \) is the mud shear strength. 

% Since our model, the modeled $\hat{f}_y$ equals the yielding force upon the terrain starts to yield (Fig. \ref{Fig. sensingFig1} B xxx curve), which results in some inaccuracy at the beginning of stance phase. 

Fig. \ref{Fig. sensing_features}B, E shows the proprioceptively measured $f_x$ by the robotic flipper during the stance phase, and the estimated mud shear strength, $k_s$, respectively. 
Similar to $k_p$, the estimated $k_s$ was inversely related to the water content, which is consistent with previous literature~\cite{liu2023adaptation}, and the ground truth measurements from the load cell (Fig. \ref{Fig. sensing_features}H).
Using the estimated $k_s$, we plotted the modeled $\hat{f_x}$ in Fig. \ref{Fig. sensingFig1}E (red dashed line in green shaded region), which aligns well with the measured force, $f_x$ (Fig. \ref{Fig. sensingFig1}E, blue curve).

% In our model, we also assume the $\hat{f}_y$ keeps the same during the holding phase, and in the real scenario, $f_y$ decreases a bit due to the residual effect~\citep{}. {\SM just found, what paper do we cite?}

\subsubsection{Extraction resistance} \label{sec:ke}

%Interestingly, the flipper not only experiences a positive $f_z$ during the insertion phase but also a negative resistance force 
During the extraction phase (Fig. \ref{Fig. sensingFig1}, orange shaded area), the proprioceptively-measured $f_z$ was in the $-z$ direction, indicating a suction force on the flipper due to mud cohesion. As the flipper begins to extract from the largest insertion depth, $z_i$, the magnitude of $f_z$ rapidly increases to a peak value, $f_e$, then steadily decreases as the flipper continues to pull up from the mud (Fig. \ref{Fig. sensingFig1}D).

This observed force profile is consistent with measurements from previous literature~\citep{chen2024reduced}, which has shown that, different from dry, cohesionless sand, the presence of fine particles (clay), when activated with water, creates attractive forces between particles and increases cohesion. Due to increased cohesion, the extraction resistance force, \(f_z \), was no longer negligible like in dry sand. The penetration resistance, during the extraction phase, $k_e$, quantifies this cohesion effect. Here we estimate $k_e$ from the proprioceptive measurements.  
%The maximum absolute value of this force (Fig. \ref{Fig. sensingFig1}D, red dot), \( f_{e} \), referred to as the extraction necking force, is proportional to the initial insertion depth, \( z_i \), and the projected surface area, \( A \), as well as the extraction necking resistance, $k_e$~\citep{chen2024reduced}. 

Similar to $k_p$, $k_e$ could be estimated by minimizing the RMSE between the modeled $\hat{f_e}(k_e)$ using Eqn. \ref{eqn1} and actual measurements, $f_z$, during the extraction phase (Fig. \ref{Fig. sensingFig1}D, between two dash vertical lines). 

%based on the \( z_i \) and $A$ at the moment the flipper starts to extract. 

% \begin{equation}
%    k_e = \frac{f_e}{ \int A(z) \, dz }
%     \label{eqn2}
% \end{equation}

% where \( f_{e} \) is modeled as the integral of the force exerted on each infinitesimal segment, \( df_{e} \) at depth \( z \), with \( df_{e} = k_{e} \cdot A(z) \cdot dz \).

Fig. \ref{Fig. sensing_features}C, F shows the proprioceptively measured $f_z$ by the robotic flipper during the extraction phase, and the estimated extraction resistance, $k_e$, respectively. 
The estimated $k_e$ exhibited a good agreement with the ground truth measurements from the load cell (Fig. \ref{Fig. sensing_features}I).
Using the estimated $k_e$, we plotted the modeled $\hat{f_e}$ in Fig. \ref{Fig. sensingFig1}D (dark blue dashed line), which aligns well with the measured force, $f_z$ (Fig. \ref{Fig. sensingFig1}E, green curve).

\begin{figure}[!b]
    \centering
    \includegraphics[width=1\linewidth]{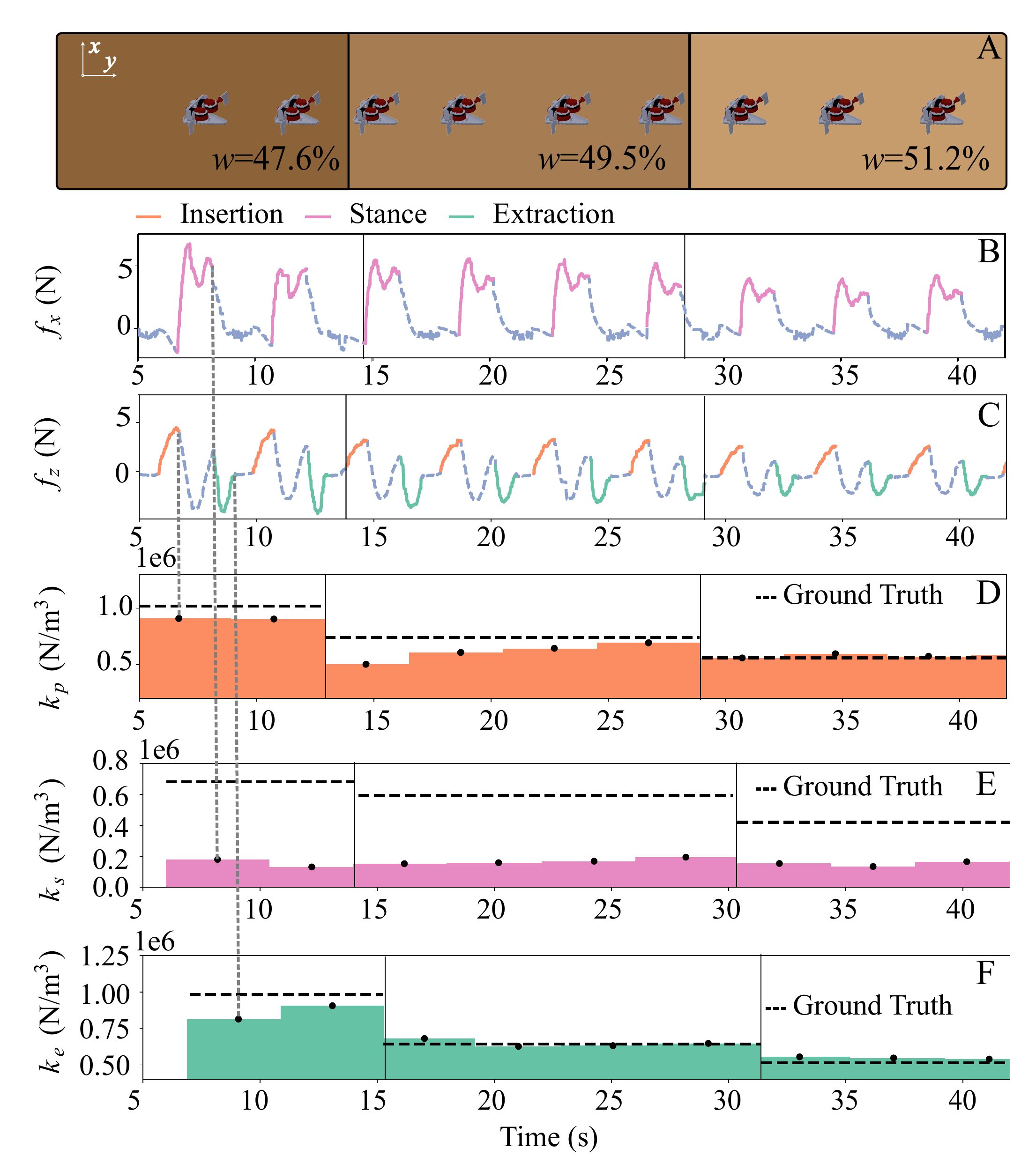}
    \caption{Mud property estimation during robot locomotion. (A) The robot moves through 3 different mud mixtures with water content \( 47.6\% \), \( 49.5\% \), and \( 51.2\% \). (B)(C) The estimated $f_x$ and $f_z$ from the right flipper over time. (D)(E)(F) The estimated $k_p$, $k_s$, and $k_e$, over time. The horizontal dashed line represents the ground truth measured using the load cell. The vertical lines represent the separation between different mixtures.}
    \label{Fig. sensingduringlocomotion}
\end{figure}

\subsection{Online mud property estimation during robot locomotion}\label{sec:online-sensing-results}
% \colorbox{yellow}{Shipeng +1}
In this section, we extend the validated 
proprioception-based mud sensing method from the static single flipper setup to a locomoting robot. 
We seek to characterize the accuracy of mud property estimation in the new setting, as the robot is no longer stationary mounted, which could induce challenges in state estimation, and more complex robot-terrain interactions. 
%and can move horizontally. Additionally, inaccuracies may also arise from the terrain's state. Specifically, when the robot moves forward, the terrain under the flipper often behaves solid-like and does not yield, causing the flipper to measure body resistance forces instead of the terrain's yield strength. 
%\subsubsection{Mud surface detection}
A key difference from the static single flipper setting is that, as the robot moves through mud, the flipper and body can disturb the mud surfaces, leading to challenges in correctly estimating penetration and shear depth of the flippers. 
To address this challenge, we detected the mud surface at each step using the proprioceptively-measured vertical force, \( f_z \). Specifically, the mud surface height was determined when \( f_z \) exceeded a force threshold of 0.5 N, which was selected based on the noise level in the force measurements.

Fig. \ref{Fig. sensingduringlocomotion}D-F shows the estimated mud properties, $k_p$, $k_s$, and $k_e$, as the robot traversed through transitioning mud mixtures with water contents of 47.6\%, 49.5\%, and 51.2\% (Fig. \ref{Fig. sensingduringlocomotion}A). Proprioceptively-measured vertical force, \( f_z \), during the penetration and the extraction phases (Fig. \ref{Fig. sensingduringlocomotion}C, orange and green, respectively) were qualitatively similar to those from single flipper experiments.
 From the robot-measured $f_z$, the estimated \( k_p \) and $k_e$ exhibits a clear distinction between three mud mixtures with varying strength (Fig. \ref{Fig. sensingduringlocomotion}D, F), closely matching the ground truth measured from the load cell (Fig. \ref{Fig. sensingduringlocomotion}D, F, black line), illustrating the potential of the proposed method to enable online mud property sensing during continuous robot locomotion. 

  %compared to using the commanded insertion depth ({\SL do we need to show the results of using commanded insertion depth}). 

% However, it is slightly higher than the ground truth (Fig. \ref{Fig. sensingduringlocomotion}F, black line). Based on our observations, this discrepancy could be attributed to the accumulation of mud mixture on the flipper during locomotion.
\begin{figure}[htbp]
    \centering
    \includegraphics[width=1\linewidth]{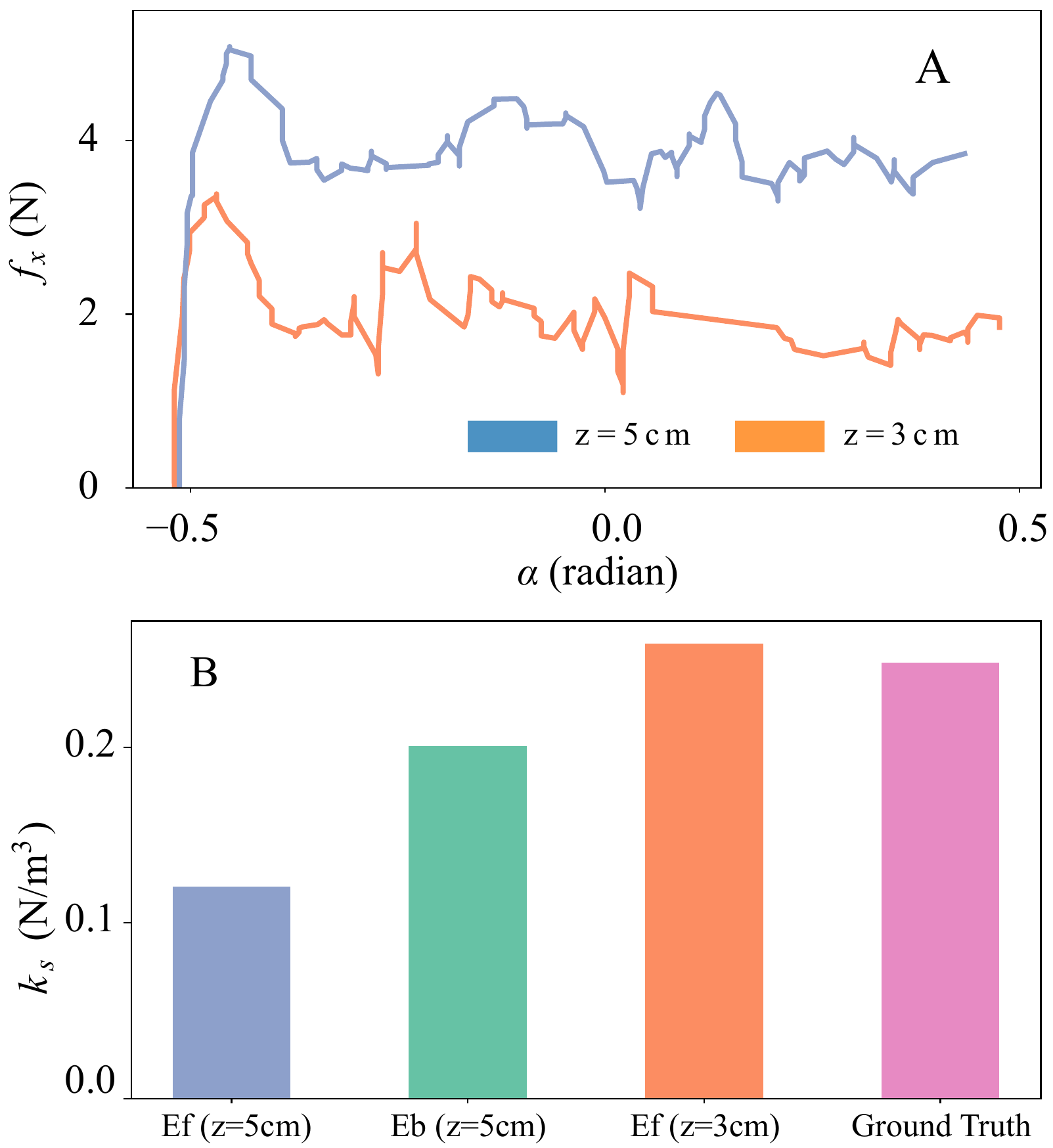}
    \caption{Proprioceptively measured horizontal force, $f_x$, and estimated mud shear strength, $k_s$, from a locomoting step from mud mixture with water content $53.8\%$. (A) The proprioceptively-measured $f_x$ during the shear phase, for flipper insertion depth, 5 cm (blue) and 3 cm (orange). (B) Estimated $k_s$ using different methods: Ef ($z$ = 5 cm) and Ef ($z$ = 3 cm) without considering body motion and mud response, Eb ($z$ = 5 cm) with compensated body drag, and ground truth from load cell measurement.}
    \label{Fig. soft_terrin_sensing_Ks}
\end{figure}

%\subsubsection{Extracting $k_s$ from body resistance v.s. from flipper resistance matters}
Interestingly, 
%even assuming accurate ground plane identification and state estimation, 
the estimated \( k_s \) exhibited a significant underestimation as compared to the ground truth (Fig. \ref{Fig. sensingduringlocomotion}E).
We hypothesized that this behavior arises from the rheology of yield‑stress materials such as sand and mud. When the applied force exceeds the yield stress~\cite{nedderman1992statics}, the material would yield and flow, whereas when the applied force remains below the yield force threshold, the material would behave like a solid.
In the robotic flipper case, the mud was forced to yield consistently throughout the stance phase. However, in a locomoting robot, the flipper initially shears through the mud, but as soon as the increasing shear force was sufficient to overcome the body drag (approximately 2.5N in our experiment), the mud solidifies, and the proprioceptively measured force no longer reflects the mud's shear strength, resulting in the observed underestimation.

% \( f_x \) during locomotion (Fig. \ref{Fig. sensingduringlocomotion}B, solid purple curve) exhibits a distinct pattern compared to the single flipper experiments. Instead of remaining constant during the latter part of the stance phase, \( f_x \) reaches a maximum at the beginning of the sweeping phase and then decreases.

To test our hypothesis, we performed the online mud sensing experiments using two different insertion depths, at 3 cm and 5 cm, respectively. The mud yield stress at 3 cm was not sufficient to overcome the body drag, and therefore the flipper continuously sheared through the fluid-like mud, resulting in an estimated $k_s$ (Fig. \ref{Fig. soft_terrin_sensing_Ks}B, orange) close to the ground truth measured by the load cell (Fig. \ref{Fig. soft_terrin_sensing_Ks}B, pink). However, due to the large flipper slippage, the robot's locomotion speed was low. 
On the contrary, the mud yield stress at 5 cm was larger than the body drag, resulting in solidified mud and an underestimated $k_s$ (Fig. \ref{Fig. soft_terrin_sensing_Ks}B, blue) that was only half of the ground truth value (Fig. \ref{Fig. soft_terrin_sensing_Ks}B, pink). The benefit was that the robot flipper could push against the solidified mud and effectively advance its body forward. 
These measurements supported our hypothesized cause of sensing discrepancy, illustrating the importance of considering substrate responses and robot body motion jointly when resolving proprioception-based force signals.
%To address the mismatch in \( k_s \) estimation during locomotion, we propose re-estimating \( k_s \) assuming the measurements are the robot's body resistance when the terrain does not yield. The robot's body resistance is generated by a drag pad, measuring 1.5 cm by 1.5 cm, mounted 2 cm below the terrain surface, and the force required for the robot's movement. To calculate the resistance force generated by the drag pad, we ran the robot without the drag pad attached and measured the thrust force required for the robot's movement, which is approximately 2.5 N. 
To account for the force discrepancy, we compensated the proprioceptively-measured $f_x$ with the body drag, to estimate  \( k_s \) when body advancement was detected (Fig. \ref{Fig. soft_terrin_sensing_Ks}B, green), which exhibited a better match as compared to the original estimated $k_s$ without account for the body drag (Fig. \ref{Fig. soft_terrin_sensing_Ks}B, blue).

\subsection{Locomotion adaptation based on online-sensed mud strength}\label{sec:loco_adapt}
% \label{sec:online-sensing-results}
% \colorbox{yellow}{Shipeng +1}

In this section, we demonstrate that the online-sensed mud strength can enable a flipper-based robot to adapt its gait parameters to mitigate catastrophic locomotion failures. 
A previous study~\citep{liu2023adaptation} discovered that flipper-driven locomotion on mud can encounter two primary failures: (1) a \textbf{flipper slippage failure} during the stance phase, that often occurs on low shear strength mud, and (2) a \textbf{flipper extraction failure} during the extraction phase, which often occurs on high extraction strength mud. 
Based on these findings and the new proprioceptive sensing method developed in this paper, we develop a sensory-based gait adaptation system, comprising an online sensing component and a gait adaptation component. Both components update once a stride cycle during the swing phase. The online sensing component uses the proposed method in Sec. \ref{sec:online-sensing-results} to estimate the mud parameters, $k_p$, $k_s$, and $k_e$, based on the state history, including the instantaneous forces, $f_x$ and $f_z$, flipper sweeping angle, $\alpha$, as well as the insertion depth, $z$.  
The gait adaptation component takes the terrain coefficients to optimize the insertion depth, $z$, to avoid the two primary locomotion failures:  %The gait controller computes the position of each motor based on inverse kinematics and the optimized insertion depth, $z$. A joint PD controller is further used to compute torque, $\tau$ for each motor. 

%Specifically, the gait adapter selects the flipper insertion depth, $z$,  to avoid these locomotion failures. 

\begin{itemize}
    \item \textbf{Flipper slippage failure:} The flipper insertion depth was chosen to satisfy the following condition to ensure that on low shear strength mud, the force generated by the flippers ($2f_x$) is larger than the body drag and inertial forces ($f_r + f_a$) that the robot needs to overcome to move forward: %terrain generated thrust force response on x direction is larger than resistance force when sweeping angle equals half of the commanded sweeping angle range~\citep{liu2023adaptation}.
    \[
    2f_x(z, k_s) > f_r(k_s) + f_a
    \]
    Here $2f_x(z, k_s)$ denotes the mud shear resistive force on two flippers, which scales with both the insertion depth, $z$, and the mud shear strength, $k_s$; $f_r$ denotes the shear resistive force on the body; and $f_a$ denotes the acceleration induced inertial force. %The sum of $f_r$ and $f_a$ represents the applied force. 

    \item \textbf{Flipper extraction failure:} To guarantee that the robot flipper can successfully extract from the mud during the extraction phase, the insertion depth should satisfy $f_{e}(z, k_e) < f_m$, where $f_{e}$ is the extraction resistive force, and $f_m$ is the maximal lifting force that the flipper could apply.
\end{itemize}

% \begin{figure}[bhtp!]
%     \centering
%     \includegraphics[width=1\linewidth]{Figs/fig7.pdf}
%     \caption{Sensory-based gait adaptation system overview.} \label{fig:system}
% \end{figure}

%The terrain model estimator calculates the terrain coefficients, $k_p$, $k_s$, and $k_e$ based on the state history, including the instantaneous forces, $f_x$ and $f_z$, sweeping angle, $\alpha$ as well as the insertion depth, $z$ online. 
% how to state we use k_p to estimate k_s simplify the logic

\begin{figure}[htbp]
    \centering
    \includegraphics[width=1\linewidth]{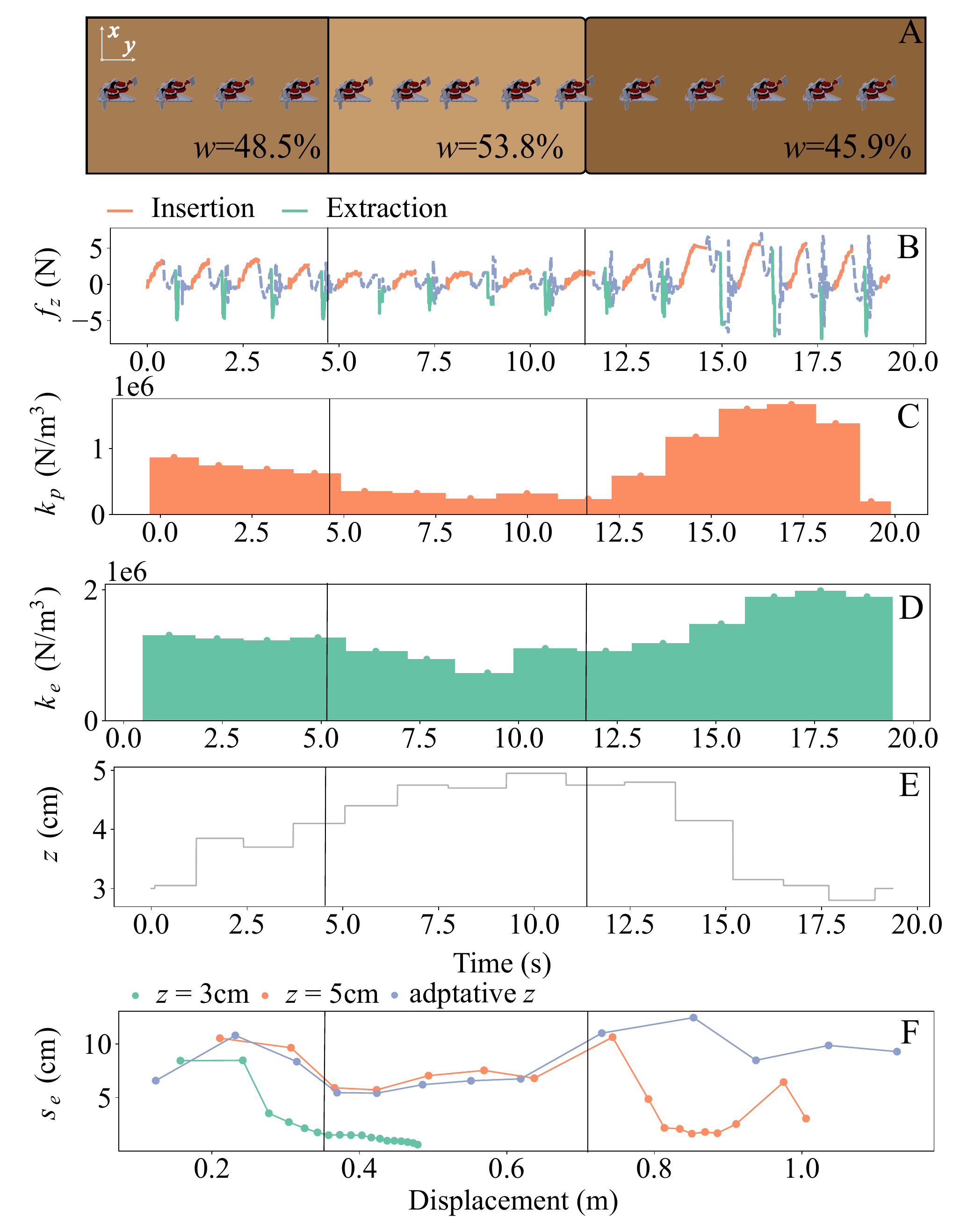}
    \caption{Sensory-based gait adaptation. (A) Adaptation experiment setup on three different mud mixtures, with water content $w=48.5\%$, $w=53.8\%$, and $w=45.9\%$. (B) The proprioceptively-measured $f_z$ over time. (C) The proprioceptively-estimated $k_p$, over time. (D) The proprioceptively-estimated $k_e$ over time. All data plotted are measurements from the right-side flipper. (E) Adaptive insertion depth, $z$, based on estimated mud properties. (F) Robot step length, $s_e$, along the $y$ direction, measured from one representative trial for each of the three gait parameters: a constant insertion depth of $z$ = 3 cm; a constant insertion depth of $z$ = 5 cm; and adaptive insertion depth shown in E, shown in blue, green, and orange, respectively. }
    \label{Fig. adaptationresult}
\end{figure}
%\subsubsection{Adaptive crutching gait with proprioceptive sensing }

% \subsubsection{Experiment setup}
% We conducted sensory-based locomotion experiments, similar to the sensing experiments, but with the robot adapting its gait based on sensory feedback during each cycle. The robot was tested in three mud mixtures with varying water content: the left segment had intermediate water content, where failures were unlikely; the middle had high water content, lacking sufficient yield stress for forward movement; and the right had low water content, making the mixture highly cohesive, preventing flipper extraction if the insertion depth was too large.

% To evaluate the performance of the gait adapter, we tested three configurations: (1) real-time insertion depth adaptation, (2) constant 3 cm insertion depth, and (3) constant 5 cm insertion depth. Each configuration was tested in 3 trials, with insertion depths and all other data recorded.

%\subsubsection{Evaluation}
To evaluate the effectiveness of the proposed sensory-based adaptation, we measured the robot stride length (\ie displacement per gait cycle) as it traverses mud mixtures with varying water content (Fig. \ref{Fig. adaptationresult}A) with three different gait parameters: (1) a constant flipper insertion depth of 3 cm (Fig. \ref{Fig. adaptationresult}F, green); (2) a constant flipper insertion depth of 5 cm (Fig. \ref{Fig. adaptationresult}F, orange); and (3) an adaptive flipper insertion depth based on the sensed mud properties (Fig. \ref{Fig. adaptationresult}F, blue).   
%mainly focus on two aspects: if the online sensing is still robust when the robot is actively changing the gait parameters and if the adaptation is effective based on the estimated terrain coefficients. 

The estimated force (Fig. \ref{Fig. adaptationresult}B) and coefficients (Fig. \ref{Fig. adaptationresult}C, D) exhibits distinct patterns for the three mud mixture regions, corresponding to the expected mud strength: lowest penetration and extraction resistance for $w = 53.8\%$, and highest penetration and extraction for $w = 45.9\%$. 
% During gait adaptation trials, we show that our proposed sensing algorithm can robustly differentiate mud mixtures with different water contents. 
Analysis of robot stride length showed that the proprioception based adaptation allowed the robot to maintain a large step length across all three mud mixtures (Fig. \ref{Fig. adaptationresult}F, blue), with a trial-wise average speed of 6.2 $\pm$ 0.3 cm/s across all three mud mixtures (Tab. \ref{table:1}). In contrast, the gait with a 3 cm insertion depth encountered a slippage failure in the low-$k_p$ mud mixture with $w = 53.8\%$ (Fig. \ref{Fig. adaptationresult}B, green), registering a speed of 2.1 $\pm$ 0.9 cm/s (Tab. \ref{table:1}). With a 5 cm insertion depth, the robot was able to move through low-$k_p$ mud, but struggled with extraction failures in high-$k_e$ mud mixture with $w = 45.9\%$ (Fig. \ref{Fig. adaptationresult}B, orange), registering a speed of 3.0 ± 1.0 cm/s (Tab. \ref{table:1}). 
These measurements demonstrated that the proprioceptively-sensed substrate properties are crucial to enabling robot gait adaptation and avoiding catastrophic locomotion failures. % that the non-adaptive locomotion algorithm experienced. This indicates the accuracy of the terrain estimation and the effectiveness of our online adaptation algorithm.  

\begin{table}[ht]
    \centering
    % Adjust row height with \arraystretch
    \renewcommand{\arraystretch}{2.3} % Increase row height
    \begin{tabular}{ c| c c c } 
    \hline
    % Header row
    & \textbf{$w=48.5\%$} & \textbf{$w=53.1\%$} & \textbf{$w=45.3\%$} \\ 
    \hline
    % Row 1: z = 3
    \multirow{1}{5em}{\centering {\textbf{\( v \, (\mathrm{cm/s}) \)} \\ \text{\( z = 3 \)}}} & 7.6$\pm$0.5  & 2.1$\pm$0.8 & 7.2$\pm$0.5 \\
    \hline
    % Row 2: z = 5
     \multirow{1}{5em}{\centering {\textbf{\( v \, (\mathrm{cm/s}) \)} \\ \text{\( z = 5 \)}}} & 6.6$\pm$0.3 & 5.5$\pm$0.7 & 3.0$\pm$1.0 \\ 
    \hline
    % Row 3: Adaptive z
    \multirow{1}{8em}{\centering \textbf{\( v \, (\mathrm{cm/s}) \)} \\ adaptive \( z \)}& 6.4$\pm$0.4 & 5.1$\pm$0.3 & 7.2$\pm$0.1 \\
    % \multirow{1}{8em}{\shortstack{\textbf{\( v \, (\mathrm{cm/s}) \)} \\ \text{ adaptive $z$}}} 
    \hline
    \end{tabular}

    % \captionsetup{justification=raggedright, singlelinecheck=false} % Align caption to the left
    % 
    \vspace{1em}
    \caption{Average forward velocity on different terrains of three different gait configurations.}

    \label{table:1}
\end{table}

\section{Limitations and Discussions}
% terrain for granular media from sand to mud, different water content. 
% how to handle imperfect protocols to gather data. 
% joint considers optimizing the sensing accuracy and locomotion trade-off. 
% \colorbox{yellow}{Shipeng +1}
A key limitation of this study is that the robot locomotion was constrained to the linear slide, which eliminates robot body pitch and roll and simplifies the challenges in state estimation and force interpretation. Future work should investigate the effectiveness of the method during unconstrained robot locomotion. According to our results, we hypothesize that for unconstrained robot on yield-stress substrates like sand and mud, the resistive force sensed by the flipper will exhibit an initial, linearly increasing force–depth response similar to the constraint case, but followed by a plateau upon mud solidification. This provides a pathway to address the limitation and generalize our methods to unconstrained robot: by coupling flipper force measurements with a simple state estimator (e.g., body lift/pitch detection via motion tracking or IMU), one could detect the yield‑to‑solidification threshold from robot state, and isolate the linear region for computing penetration resistance. This generalization will allow our methods to be applied 
%This means that, regardless of whether the robot is constrained or free to move, or whether its appendages differ in shape or size, the same approach will reliably extract substrate stiffness online. Consequently, our method should generalize
across diverse platforms and a wide variety of yield‑stress terrains, enabling robust, real‑time adaptation to challenging field conditions.

The other limitation is that the proposed method is currently specific to the bio-inspired crunching gait, where the robot flipper executes strictly horizontal and vertical movements, limiting the gait space that a robot could use. Future work will explore how the proposed method could be extended to more generalized trajectories. % parameter-based resistance force theory to allow robots to retrieve terrain information from arbitrary movements in deformable terrain. 

\section{Conclusion}
% \colorbox{yellow}{Shipeng +1}
%Online adaptability is essential for autonomous robots in field exploration missions, as they must navigate complex, deformable natural terrains 
This study presents a proprioceptive-based method for characterizing deformable terrain properties using actuator signals from robot-terrain interactions. By comparing the proprioceptive measurements with load cell data, a direct-drive actuated robotic flipper could accurately estimate penetration resistance, shear strength, and extraction resistance from mud mixtures with varying levels of strength and cohesion. 
We demonstrated that the proposed method enabled robots to detect changes in substrate properties during locomotion, without any external sensors. These estimated properties allow the robot to effectively adjust its gait in real time, preventing locomotion failures and improving mobility across transitioning muddy substrates.

Our findings highlight the importance of terrain-aware proprioceptive sensing and adaptive gait strategies in enabling robust robot mobility on complex, deformable substrates. Future research can extend this method by investigating additional gait trajectories, applying it to diverse robotic platforms, and evaluating its effectiveness on various substrates. These advancements can benefit a wide range of robotic applications, such as mudslide assessment, nearshore characterization, and extraterrestrial exploration missions. 

\section*{Acknowledgments}
This work is supported by the National Science Foundation (NSF) CAREER award \# 2240075, the NASA Planetary Science and Technology Through Analog Research (PSTAR) program, Award \# 80NSSC22K1313, the NASA Lunar Surface Technology Research (LuSTR) program, Award \# 80NSSC24K0127, and the NASA Mars Exploration Program (MEP) Technology Development Funding.
% % \clearpage

%% Use plainnat to work nicely with natbib. 

\bibliographystyle{plainnat}
\bibliography{main}

\end{document}